\documentclass[runningheads, envcountsame, a4paper]{llncs}
\usepackage{subfigure}
\usepackage[hyphens]{url}
\usepackage[misc]{ifsym}
\usepackage{array,tabularx,booktabs}
\usepackage{multirow}
 \usepackage[table,xcdraw]{xcolor}
\usepackage[pdftex]{graphicx}
\usepackage{epstopdf}
\toctitle{Automatic Pass Annotation from Soccer Video Streams Based on Object Detection and LSTM}
\tocauthor{D. Sorano et al.}
\begin{document}
\title{Automatic Pass Annotation from Soccer Video Streams Based on Object Detection and LSTM}
\titlerunning{Automatic Pass Annotation From Soccer Video Streams}

\author{Danilo Sorano\inst{1} \and
Fabio Carrara \inst{2} \and
Paolo Cintia \and  \inst{1}
\\ Fabrizio Falchi \inst{2} \and Luca Pappalardo \inst{2} \Letter}

\authorrunning{D. Sorano et al.}

\institute{Department of Computer Science, University of Pisa, Italy \and
ISTI-CNR, Pisa, Italy \\ 
\email{luca.pappalardo@isti.cnr.it}
}

\maketitle              
\begin{abstract}
Soccer analytics is attracting increasing interest in academia and industry, thanks to the availability of data that describe all the spatio-temporal events that occur in each match. 
These events (e.g., passes, shots, fouls) are collected by human operators manually, constituting a considerable cost for data providers in terms of time and economic resources.
In this paper, we describe PassNet, a method to recognize the most frequent events in soccer, i.e., passes,  from video streams.
Our model combines a set of artificial neural networks that perform feature extraction from video streams, object detection to identify the positions of the ball and the players, and classification of frame sequences as passes or not passes. 
We test PassNet on different scenarios, depending on the similarity of conditions to the match used for training. Our results show good classification results and significant improvement in the accuracy of pass detection with respect to baseline classifiers, even when the match's video conditions of the test and training sets are considerably different. PassNet is the first step towards an automated event annotation system that may break the time and the costs for event annotation, enabling data collections for minor and non-professional divisions, youth leagues and, in general, competitions whose matches are not currently annotated by data providers.

\keywords{Sports Analytics  \and Computer Vision \and Applied Data Science \and Deep Learning \and Video Semantics Analysis.}
\end{abstract}
\section{Introduction}

Soccer analytics is developing nowadays in a rapid way, thanks to sensing technologies that provide high-fidelity data streams extracted from every match and training session \cite{gudmundsson2017spatio,pappalardo_cintia_2019,rossi2018effective}. 
In particular, the combination of video-tracking data and soccer-logs, which describe the movements of players and the spatio-temporal events that occur during a match, respectively, allows sophisticated technical-tactical analyses \cite{wei2013large,fernandez2018wide,pappalardo2017quantifying,pappalardo2019playerank,cintia2015harsh,decroos2019actions}. 
However, from a data provider's perspective, the collection of soccer-logs is expensive, time-consuming, and not free from errors \cite{liu2013inter}. It is indeed still performed manually through proprietary software for the annotation of events (e.g., passes, shots, fouls) from video streams, a procedure that requires around three human operators and about two hours per match \cite{pappalardo_cintia_2019}. 
Given these costs and the enormous number of matches played every day around the world, data providers collect data regarding relevant professional competitions only, of which they sell data to clubs, companies, websites, and TV shows.
For all these reasons, an automated event annotation system would provide many benefits to the sports industry. 
On the one hand, it would bring a reduction of errors, time, and costs of annotation for data providers:
an automatic annotation system may substitute one of the human operators, or it may be used to check the reliability of the events collected manually.
On the other hand, it would enable data collections for non-professional divisions, youth leagues and, in general, competitions whose matches data providers have no economic convenience to annotate.

Most of the works in the literature focus on video summarization, i.e., the detection from video streams of salient but infrequent episodes in matches, such as goals, replays, highlights, and play-breaks \cite{bayat2014goal,jiang2016automatic,saraogi2016event,liu2017soccer,yu2019soccer}. Another strand of research focuses on the identification of players through their jersey number \cite{gerke2015soccer} or on the detection of ball possession \cite{khan2018soccer}.
Nevertheless, there are no approaches that focus specifically on the recognition of \emph{passes} from video streams, although passes correspond to around 50\% of all the events in a soccer match \cite{pappalardo_cintia_2019}. 
It hence goes without saying that a system that wants to drastically reduce errors, time and economic resources required by event annotation must be able to accurately recognize passes from video streams.

In this paper, we propose {\sf PassNet}, a computer vision system to detect passes from video streams of soccer matches. 
We define a \emph{pass} as a game episode in which a player kicks the ball towards a teammate, and we define \emph{automatic pass detection} as the problem of detecting all sequences of video frames in which a pass occurs. 
{\sf PassNet} solves pass detection combining three models: ResNet18 for feature extraction, a Bidirectional-LSTM for sequence classification, and YOLOv3 for the detection of the position of the ball and the players. 
To train {\sf PassNet}, we integrate video streams of four official matches with data describing when each pass begins and ends, collected manually through a pass annotation application we implement specifically to this purpose.  
We empirically prove that {\sf PassNet} overtakes several baselines on different scenarios, depending on the similarity of conditions of the match's video stream used for training, and that it has good agreement with the sets of passes annotated by human operators of a leading data collection company.
Given its flexibility and modularity, {\sf PassNet} is a first step towards the construction of an automated event annotation tool for soccer.

\section{Related Work}

De Sousa et al. \cite{de2011overview} group event detection methods for soccer in low-level, medium-level, and high-level analysis. The low-level analysis concerns the recognition of basic marks on the field, such as lines, arcs, and goalmouth. The middle-level analysis aims at detecting the behavior of the ball and the players. The high-level analysis regards the recognition of events and video summarization.

\emph{Video summarization}. Most of the works in the literature focus on the detection from video streams of salient actions such as goals, replays, highlights, and play-breaks.
Bayat et al. \cite{bayat2014goal} propose a heuristic method to detect goals from video streams based on the audio intensity, color intensity, and the presence of goalmouth. Zawbaa et al. \cite{zawbaa2012event} perform video summarization through shot-type classification, play-break classification, and replay detection, while Kapela et al. \cite{kapela2014real} can detect scores and near misses that do not result in a score.
Jiang et al. \cite{jiang2016automatic} detect goals, shots, corner kicks, and cards through a combination of convolutional and recurrent neural networks that proceeds in three progressive steps:
play-break segmentation, feature extraction, and event detection.
Yu et al. \cite{yu2019soccer} use deep learning to identify replays and associated events such as goals, corners, fouls, shots, free-kicks, offsides, and cards.
Saraogi et al. \cite{saraogi2016event} develop a method to recognize notable events combining generic event recognition, event's active region recognition, and shot classification. Liu et al. \cite{liu2017soccer} use 3D convolutional networks to perform play-break segmentation and action detection from video streams segmented with shot boundary detection.  Similarly, Fakhar et al. \cite{fakhar2019event} address the problem of highlight detection in video streams in three steps: shot boundary detection, shot view classification, and replay detection. 

\emph{Ball, player and motion detection.}
Kahn et al. \cite{khan2018soccer} use object detection methods based on deep learning to identify when a player is in possession of the ball.  
Gerke et al. \cite{gerke2015soccer} use a convolutional neural network to recognize the jerseys from labeled images of soccer players. 
Carrara et al. \cite{carrara2019lstm} use recurrent neural networks to annotate human motion streams, in which each step represents 31 joints of a skeleton, each described as a point in a 3D space.

\emph{Contribution of our work.}
An overview of the state of the art cannot avoid noticing that there are no approaches that focus on the recognition of \emph{passes} from video streams. Nevertheless, automatic pass detection is essential, considering that passes correspond to around 50\% of all the events in a soccer match \cite{pappalardo_cintia_2019}. In this paper, we fill this gap by providing a method, based on a combination of artificial neural networks, that can recognize passes from video streams. 
\section{Pass detection problem}
\label{sec:pass_detection}
An \emph{event} is any relevant episode that occurs at some point in a soccer match, e.g., pass, shot, goal, foul, save. 
The type of events annotated from video streams is similar across different data collection companies, although there may be differences in the way annotated events are structurally organized  \cite{liu2013inter,pappalardo_cintia_2019}.
From a video stream's perspective, an event is the sequence of $n$ frames $\langle k_t, k_{t + 1}, \dots, k_{t + n} \rangle$ in which it takes place.


Nowadays, events are annotated from video streams through a manual procedure performed through a proprietary software (the tagger) by expert video analysts (the operators) \cite{pappalardo_cintia_2019}. For example, the company Wyscout\footnote{\url{https://wyscout.com/}} uses one operator per team and one operator acting as a responsible supervisor of the output of the whole match \cite{pappalardo_cintia_2019}.
Each operator annotates each relevant episode during the match, hence defining the event's type, sub-type, coordinates on the pitch, and additional attributes \cite{pappalardo_cintia_2019}. 
Finally, the operators perform quality control by checking the coherence between the events that involve both teams, and through manually scanning the annotated events. Manual event annotation is time-consuming and expensive: since the annotation of a single match requires about two hours and three operators, the effort and the costs needed to tag an entire match-day are considerable \cite{pappalardo_cintia_2019}.

We define automatic event detection as the problem of annotating sequences of frames in a video stream with a label representing the corresponding event that occurs during that frame sequence. 
In particular, in this paper we focus on \emph{automatic pass detection}: detecting all sequences of video frames in which a \emph{pass} occurs. A pass is a game episode in which a player in possession of the ball tries to kick it towards a teammate. 
\section{PassNet}

\begin{figure}[htb]
    \centering
    \includegraphics[width=1.0\textwidth]{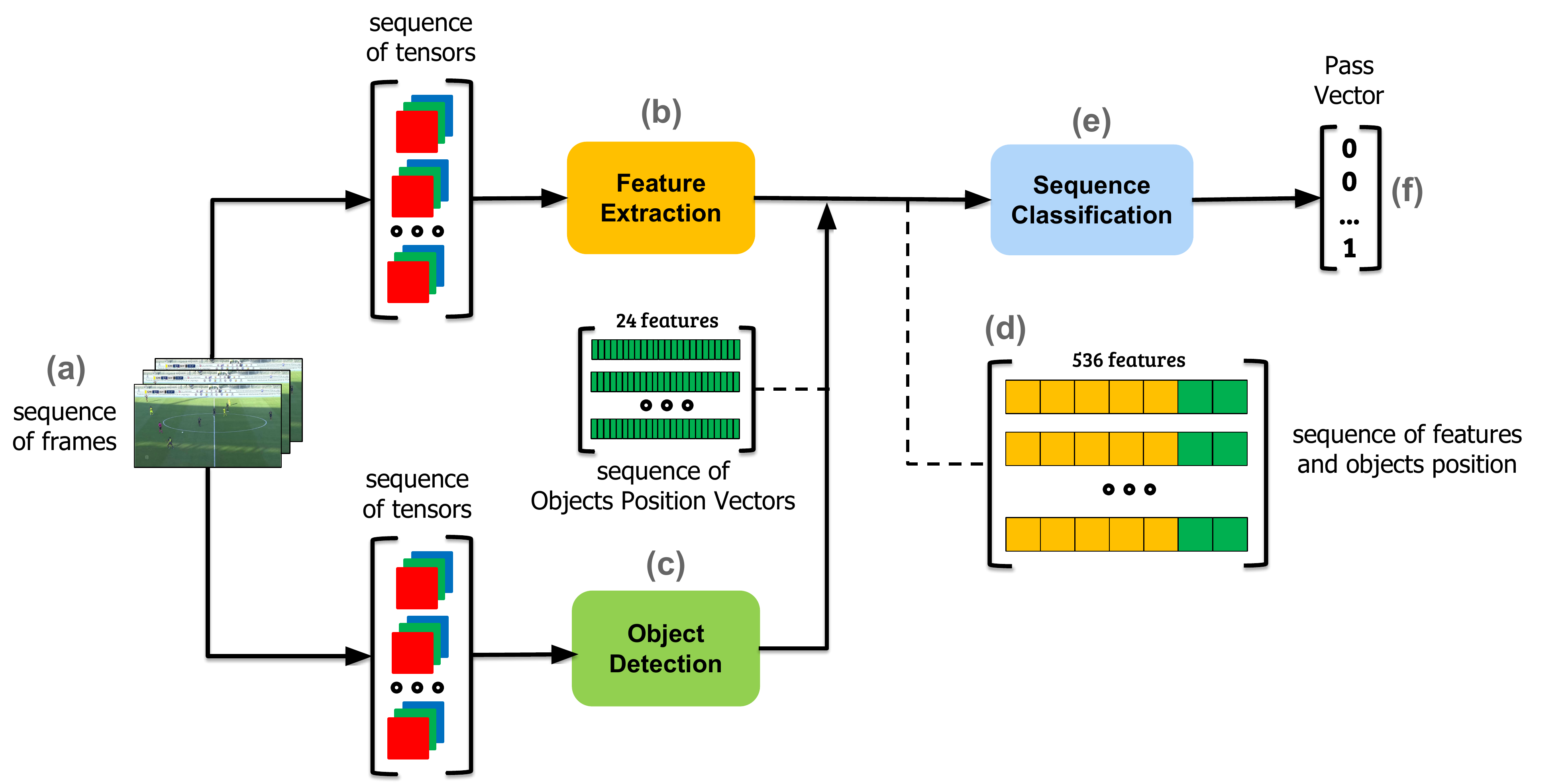}
    \caption{\textbf{Architecture of {\sf PassNet}}. The sequence of frames of the video stream (a) is provided to a Feature Extraction module (b) that outputs a sequence of vectors of 512 features, and to an Object Detection module (c) that outputs a sequence of vectors of 24 features describing the position of the ball and the closest players to it. 
    The two outputs are combined into a sequence of vectors of 536 features (d) and provided to a Sequence Classification module (e), which outputs a pass vector (f) that indicates, for each frame of the original sequence, whether or not it is part of a pass sequence.}
    \label{fig:passnet}
\end{figure}

Our solution to the automatic pass detection problem is {\sf PassNet}, the architecture of which is shown in Figure \ref{fig:passnet}.\footnote{{\sf PassNet}'s code and data are available at \url{github.com/jonpappalord/PassNet}} It combines three tasks: \emph{(i)} \emph{feature extraction} reduces the dimensionality of the input using ResNet18 (Section \ref{sec:feature_extraction}); \emph{(ii)} \emph{object detection} detects the players and the ball in the video frames using YOLOv3 (Section \ref{sec:object_detection}); \emph{(iii)} \emph{sequence classification} classifies sequences of frames as containing a pass or not using a Bi-LSTM \cite{carrara2019lstm} (Section \ref{sec:sequence_processing}). 

In {\sf PassNet}, 
each frame has dimension $3 \times 352 \times 240$, where 3 indicates the RGB channel, and $352 \times 240$ is the size of an input frame.
The sequence of frames that composes a video stream (Figure~\ref{fig:passnet}a) is provided in input to \emph{(i)} a feature extraction module (Figure~\ref{fig:passnet}b), which outputs a sequence of vectors of 512 features, and \emph{(ii)} to an object detection module (Figure~\ref{fig:passnet}c), which outputs a sequence of vectors of 24 features describing the positions of the ball and the closest players to it. 
The two outputs are combined into a sequence of vectors of 536 features (Figure \ref{fig:passnet}d) and provided as input to a sequence classification module (Figure \ref{fig:passnet}e), which generates a pass vector (Figure \ref{fig:passnet}f) that indicates, for each frame of the original sequence, whether or not it is part of a pass sequence.



\subsection{Feature Extraction}
\label{sec:feature_extraction}
The sequence of frames is provided in input to the Feature Extraction module frame by frame, each of which is transformed into a feature vector by the image classification model ResNet18 \cite{he2016deep}.
In the end, the feature vectors are combined again into a sequence.
ResNet18 consists of a sequence of convolution and pooling layers \cite{he2016deep}. 
Convolution layers use convolution operations to produce a feature map by sliding a kernel of fixed size over the input tensor and computing the dot-product between the covered input and the kernel weights. Pooling layers reduce the dimensionality of each feature map while retaining the most important information.
ResNet18 returns as output a feature vector of 1,000 elements, which represents all possible classes (objects) in the ImageNet-1K data set \cite{imagenet_data} used to train the model. For our purpose, we take the output of the last average pooling layer of ResNet18, which generates a vector of 512 features.

\subsection{Object Detection}
\label{sec:object_detection}
\begin{figure}[htb!]
    \centering
    \includegraphics[width=1.0\textwidth]{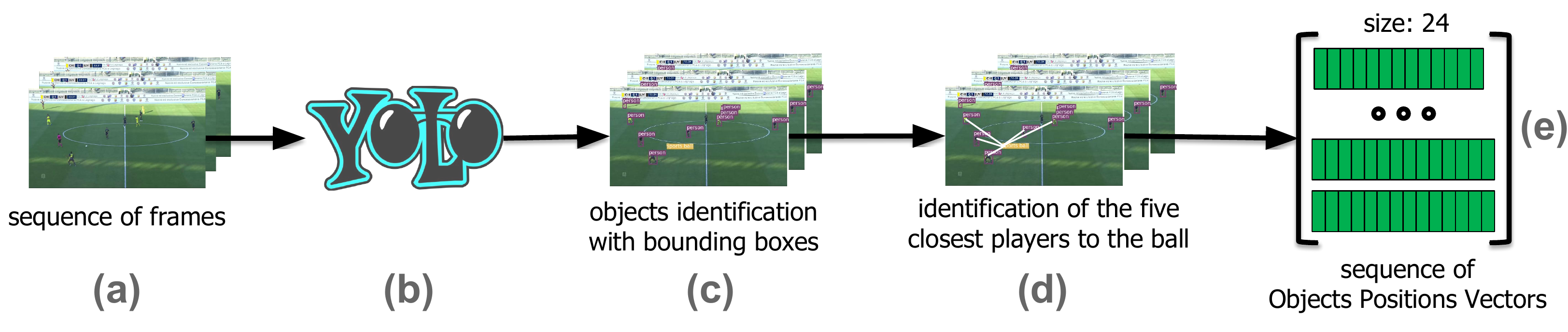}
    \caption{\textbf{Object Detection module}. The sequence of frames (a) is provided to YOLOv3 \cite{redmon2018yolov3} (b), which detects the players and the ball in each frame (c). The five closest players to the ball are identified (d) and a vector of 24 elements is created describing the positions of the objects (e).}
    \label{models:pos_info}
\end{figure}

To identify the position of relevant objects in each frame, such as the players and the ball, we use YOLOv3 \cite{redmon2018yolov3}, a convolutional neural network for real-time object detection. In particular, we use a version of YOLOv3 pre-trained on the COCO dataset \cite{lin2014coco}, which contains images labeled as balls or persons. YOLOv3 assigns to each object detected inside a frame a label and a bounding box, the center of which indicates the position of the object inside the frame. Figure \ref{models:pos_info} summarizes the process of extraction of the position of the objects using YOLOv3.

\begin{figure}[htb!]
    \centering
    \includegraphics[width=1.0\textwidth]{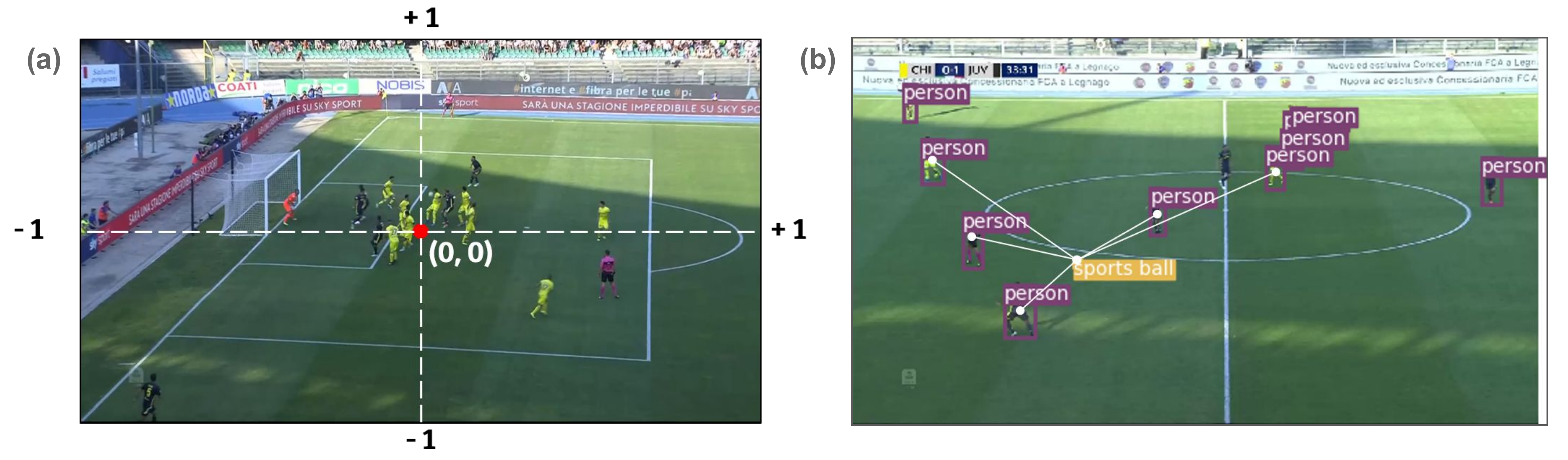}
    \caption{\textbf{Construction of the Object Position Vectors.} (a) Normalization of the coordinates in the range $[-1, +1]$, where the center of the frame has coordinates $(0,0)$. (b) The five players identified by YOLOv3 with the shortest distance from the ball.}
    \label{fig:normalization_and_players}
\end{figure}

We provide to YOLOv3 a match's video stream frame by frame. Based on the objects identified by YOLOv3, we convert each frame into a vector of 24 elements, that we call \emph{Objects Position Vector} (OPV). OPV combines six vectors of length four, describing the ball and the five closest players to the ball. In particular, each of the six vectors has the following structure:
\begin{itemize}
    \item the first element is a binary value, where 0 indicates that the vector refers to the ball and 1 that it refers to a player;
    \item the second element has value 1 if the object is detected in the frame, and 0 that the vector is a flag vector (see below);
    \item the third and the fourth elements indicate the coordinates of the object's position, normalized in the range $[-1, +1]$, where the center of the frame has coordinates $(0,0)$ (Figure \ref{fig:normalization_and_players}a).
\end{itemize}

For example, vector $[0, 1, 0.8, -0.1]$ indicates that YOLOv3 detects that ball in a frame at position $(0.8, -0.1)$, while $[1, 1, -0.1, 0.4]$ indicates that YOLOv3 detects a player at position $(-0.1, 0.4)$. Note that YOLOv3 may identify no objects in a frame, even if they are actually present \cite{redmon2018yolov3}.\footnote{In our experiments, we find that this situation happens for 0.66\% of the frames.} When an object is not detected, we substitute the corresponding vector with a \emph{flag vector}. Specifically, if in a frame the ball is not detected, we describe the ball using flag vector~$[0, 0, 0, 0]$. 

We detect the five closest players to the ball by computing the distance to it of all the players detected in a frame (Figure \ref{fig:normalization_and_players}b).
If less than five players are detected, we describe a player using the flag vector~$[1, 0, 2, 2]$. When no player or ball is detected, we use flag vectors for both the ball and the players. If at least one player is detected, but the ball is not detected, we assume that the ball is located at the center of the frame and identify the five closest players to it. 

\subsection{Sequence Classification}
\label{sec:sequence_processing}
\begin{figure}[htb]
    \centering
    \includegraphics[width=1.0\textwidth]{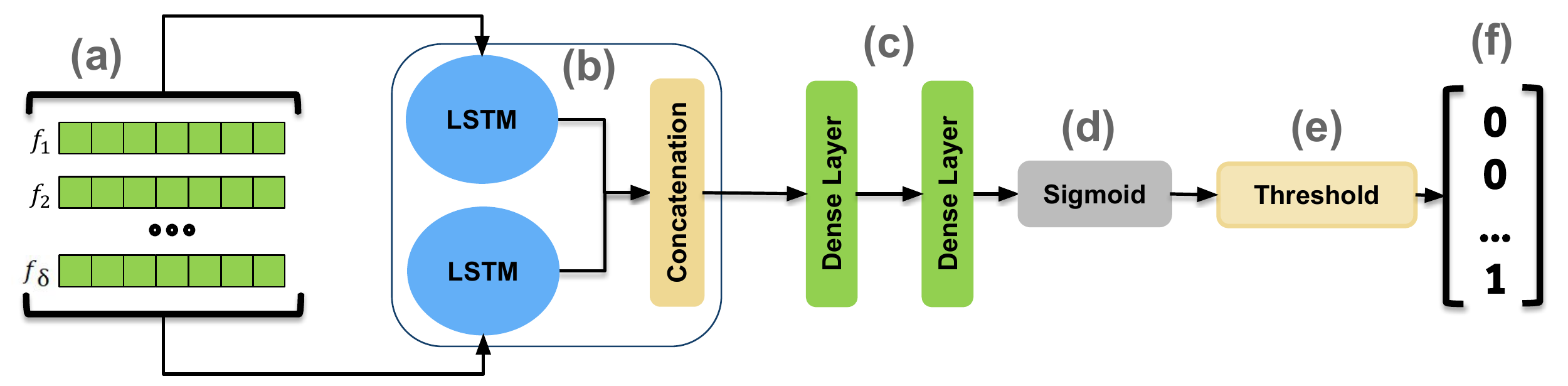}
    \caption{\textbf{Sequence Classification Module}. Each sub-sequence of $\delta$ vectors (a) is provided to a Bi-LSTM (b), which processes the sub-sequence in two directions: from the first frame $f_1$ to the last frame $f_\delta$ and from $f_\delta$ to $f_1$. The output of the Bi-LSTM goes to two dense layers (c) with ReLu activation functions and dropout=0.5. A sigmoid activation function (d) and an activation threshold (e) are used to output the pass binary vector, in which 1 indicates the presence of a pass in a frame.}
    \label{fig:sequence_classification}
\end{figure}

The outputs of the Feature Extraction module and the Object Detection module are combined into a sequence of vectors of 536 features and provided in input to the Sequence Classification module (Figure \ref{fig:sequence_classification}). 
We use a sliding window $\delta$ to split the sequence of vectors into sub-sequences of length $\delta$.
Each sub-sequence (Figure \ref{fig:sequence_classification}a) goes in input to a Bidirectional-LSTM (Figure \ref{fig:sequence_classification}b), followed by two dense layers (Figure \ref{fig:sequence_classification}c) that output a vector of $\delta$ values.
Each element of this vector is transformed into 1 (\texttt{Pass}) or 0 (\texttt{No Pass}) according to a sigmoid activation function (Figure \ref{fig:sequence_classification}d) and an activation threshold (Figure \ref{fig:sequence_classification}e). The best hyper-parameter values of the Sequence Classification module (e.g., number of hidden units in the dense layers, value of activation threshold) are determined experimentally.

\section{Data sets}
\label{sec:experiments_results}

We have video streams corresponding to four matches in the Italian first division: AS Roma vs. Juventus FC, US Sassuolo vs. FC Internazionale, AS Roma vs. SS Lazio from season 2016/2017, and AC Chievo Verona vs. Juventus FC from season 2017/2018.
All of them are video broadcasts on TV and have resolution $1280\times720$ and 25 frames per second. We eliminate the initial part of each video, in which the teams' formations are presented and the referee tosses the coin, and we split the videos into the first and second half. For computational reasons, we reduce the resolution of the video to $352 \times 240$ and 5 frames per second.

We associate each video with an external data set containing all the spatio-temporal events that occur during the match, including passes. These events are collected by Wyscout through the manual annotation procedure described in Section \ref{sec:pass_detection}.
In particular, each pass event describes the player, the position on the field, and the time when the pass occurs \cite{pappalardo_cintia_2019}.

We use the time of a pass to associate it with the corresponding frame in the video. 
Unfortunately, an event indicates the time when the pass starts, but not when it ends. 
Moreover, by comparing the video and the events, we note that the time of an event is often misaligned with the video.
We overcome these drawbacks by annotating manually the passes through an application we implement specifically to this purpose (see Section \ref{sec:pass_annotation_application}). 

After the manual annotation, for each match, we construct a vector with a length equal to the number of frames in the corresponding video. In this vector, each element can be either 1, indicating that the frame is part of a sequence describing a pass (\texttt{Pass}), or 0, indicating the absence of a pass in the frame (\texttt{No Pass}). For example, vector $[0011111000]$ indicates that there are five consecutive frames in which there is a pass.

\subsection{Pass Annotation Application} \label{sec:pass_annotation_application}
\begin{figure}[htb!]
    \centering
    \includegraphics[width=0.9\textwidth]{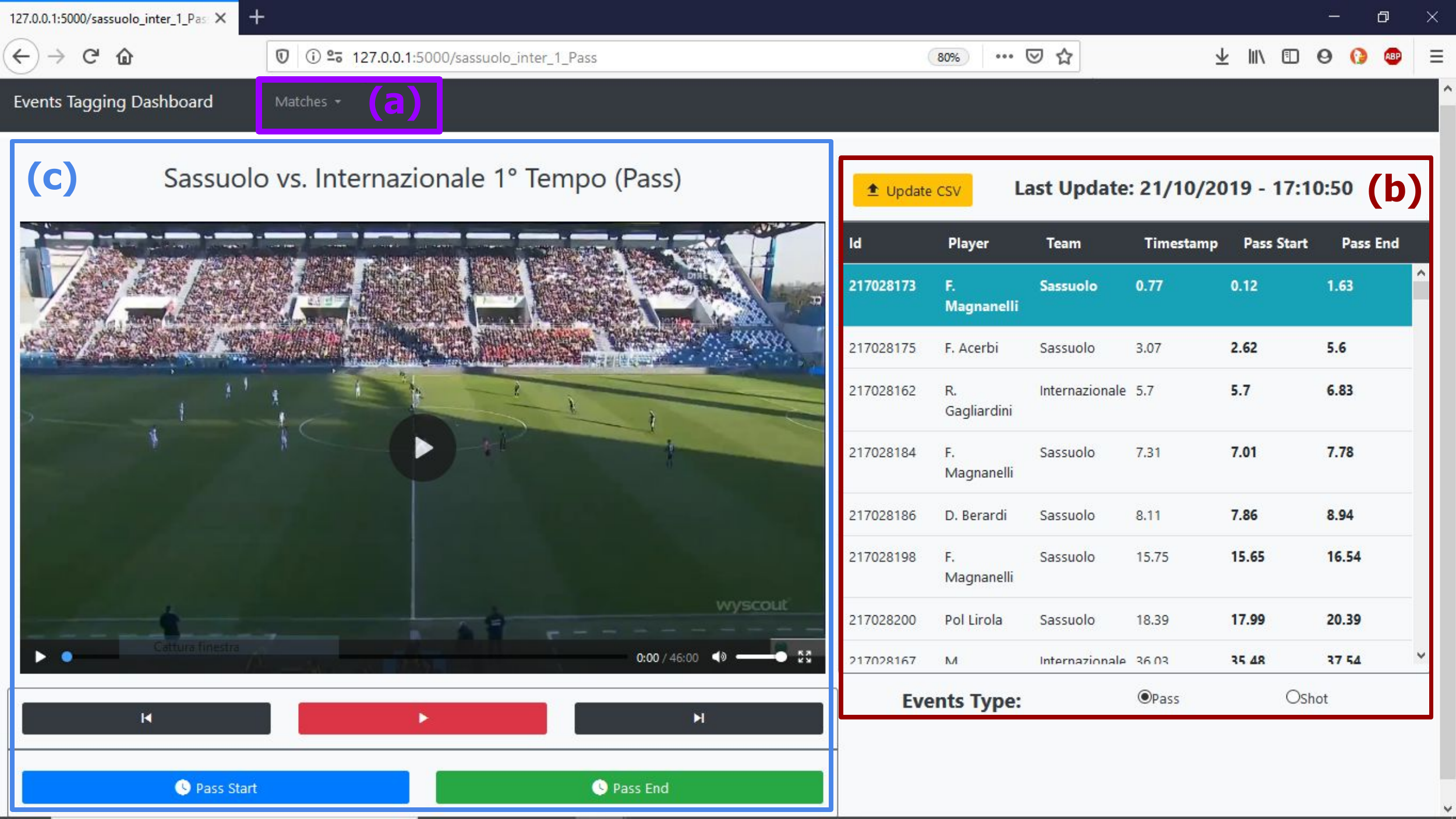}
    \caption{\textbf{Visual interface of the manual annotation application}. The user can load a match using the appropriate dropdown (a). On the right side (b), a table shows all the pass events of the match and related information. On the left side (c), the interface shows the video and buttons to start and pause it, to move backward and forward, and to annotate the starting and ending times. When the user clicks on a row in the table, the video moves to two seconds before the event. A video that illustrates the functioning of the application is here: \texttt{\url{https://youtu.be/vO98f3XuTAU}}.}
    \label{fig:application_interface}
\end{figure}

We implement a web application that contains a user interface to annotate the starting and ending times of a pass.\footnote{The application is developed using python framework Flask, and is available at \url{https://github.com/jonpappalord/PassNet}}
Figure \ref{fig:application_interface} shows the structure of the application's visual interface. 
When the user loads a match using the appropriate dropdown (Figure \ref{fig:application_interface}a), a table on the right side shows the match's passes and related information (Figure \ref{fig:application_interface}b). 
On the left side, the interface shows the video and buttons to play and pause it, to move forward and backward, and to tag the starting time (\texttt{Pass Start}) and the ending time (\texttt{Pass End}) of a pass (Figure \ref{fig:application_interface}c). When the user clicks on a row in the table to select the pass to annotate, the video automatically goes at the frame two seconds before the pass time. At this point, the user can use the \texttt{Pass Start} and \texttt{Pass End} buttons to annotate the starting and ending times, that will appear in the table expressed in seconds since the beginning of the video. In total, we annotate 3,206 passes, which are saved into a file that will be used to train {\sf PassNet} and evaluate its performance.

\section{Experiments}
We compare {\sf PassNet} with the following models:
\begin{itemize}
\item {\sf ResBi} uses just the Feature Extraction module and the Sequence Classification module, i.e., it does not use the Object Detection module for the recognition of the position of the ball and the players;

\item {\sf Random} predicts the label randomly;
\item {\sf MostFrequent} predicts always the majority class, \texttt{No Pass} (71\% of the frames);
\item {\sf LeastFrequent}  predicts always the minority class, \texttt{Pass} (29\% of the frames). 
\end{itemize}


To classify a frame as \texttt{Pass} or \texttt{No Pass} we try several activation thresholds (Figure \ref{fig:passnet}g): 0.5, 0.9, and the threshold that maximizes the Youden Index (YI) \cite{BERRAR2019546}, where $\mbox{YI} = Rec + True Negative Rate - 1$, and $\mbox{YI} \in [0, 1]$. YI represents in the model's ROC curve the farthest point from the random classifier's curve. 
We compare the models in terms of accuracy ($Acc$), F1-score ($F1$), precision on class \texttt{Pass} ($Prec$), precision on class \texttt{No Pass} ($PrecNo$), recall on class \texttt{Pass} ($Rec$), recall on class \texttt{No Pass} (RecNo) \cite{tan2016introduction}. 

We perform the experiments on a computer with CPU Intel(R) Core(TM) i7-6800K CPU @ 3.40GHz, 32 GB RAM, GPU GeForce GTX 1080.
The time required to train the model (11 epochs) on a match half is around 11 hours; the time required for the annotation of a match half is on average 30 minutes.



\subsection{Results}
We fix some  hyper-parameter values and use the first half of AS Roma vs. Juventus FC to tune the learning rate, the sequence dimension $\delta$, and the hidden dimension of the Sequence Classification module of {\sf PassNet} and {\sf ResBi} (Table \ref{tab:all_thresholds}).
We use each hyper-parameter configuration to train {\sf PassNet} and {\sf ResBi} on a validation set extracted from the training set and test the configuration on the second half of AS Roma vs. Juventus FC. 
In particular, we try values 128, 256 and 512 for the hidden dimension, $\delta=10, 25$, and values 0.01, 0.001 and 0.0001 for the learning rate.
We evaluate the performance of each configuration in terms of average precision ($AP$), the weighted mean of precision at several thresholds. 
Formally, $AP = \sum_{n} (Rec_n - Rec_{n-1})Prec_n$, where $Prec_n$ and $Rec_n$ are precision and recall at the $n$-th threshold, respectively, and $Rec_{n-1}$ is the recall at the $(n-1)$-th threshold \cite{tan2016introduction}.
Table \ref{tab:all_thresholds} shows the hyper-parameter values corresponding to the best configuration of {\sf PassNet} and {\sf ResBi}.

We test {\sf PassNet}, {\sf ResBi} and the baselines on four scenarios, in which we use: 
\emph{(i)} the same match used for training the model (\texttt{Same} scenario); 
\emph{(ii)} a match that shares similar video conditions as the match used for training the model (\texttt{Similar} scenario); 
\emph{(iii)} matches with different teams and light conditions (\texttt{Different} scenario); \emph{(iv)} a mix of matches with similar and different conditions (\texttt{Mixed} scenario). 

\begin{table}[]
\newcolumntype{Y}{>{\centering\arraybackslash}X}
\begin{tabularx}{\linewidth}{*{12}{Y}}
\toprule
 & \multicolumn{7}{c}{\bf fixed hyper-parameters}           & \multicolumn{4}{c}{\bf \scriptsize tuned hyper-parameters}        \\
\cmidrule(l){2-8}
\cmidrule(l){9-12}
 & input dim & yolo dim & layer dim & dense dim & drop out & batch size & opti- mizer & hidden dim & seq dim & learn. rate & best epoch \\
\midrule
\sf PassNet & 512       & 24       & 1         & 2           & 0.5     & 1          & adam      & \bf 128        & \bf 25          & \bf 0.0001        & \bf 6 \\
\sf ResBi & 512       & -       & 1         & 2           & 0.5     & 1          & adam      & \bf 256        & \bf 25          & \bf 0.0001        & \bf 4 \\
\bottomrule
\end{tabularx}
\caption{Hyper-parameter values of the best configuration of {\sf PassNet} (AP=74\%) and {\sf ResBi} (AP=75\%). Tuning performed using the first half of AS Roma vs. Juventus FC.}
\label{tab:all_thresholds}
\end{table}

In the \texttt{Same} scenario we use two matches: we train the models on the first half of AS Roma vs. Juventus FC and test them on the second half of the same match; similarly we train the models on the first half of US Sassuolo vs. Internazionale FC and test them on the second half of the match.
On AS Roma vs. Juventus FC, {\sf PassNet} and {\sf ResBi} have similar performance in terms of F1-score ($F1_{\mbox{\scriptsize PassNet}}=70.21\%$, $F1_{\mbox{\scriptsize ResBi}}=70.50\%$), and they both outperform the baseline classifiers with an improvement of 21.24\% absolute and 43\% relative with respect to the best baseline, \texttt{LeastFrequent} ($F1_{\mbox{\scriptsize Least}}=49.26\%$, see Table \ref{tab:same_scenario_comparison}).  
 On US Sassuolo vs. FC Internazionale, {\sf PassNet} has lower performance than the other match but still outperforms {\sf ResBi} and the baselines ($F1_{\mbox{\scriptsize PassNet}}=54.44\%$, $F1_{\mbox{\scriptsize ResBi}}=53.72\%$), with an improvement of 15.75\% absolute and 41\% relative with respect to \texttt{LeastFrequent} ($F1_{\mbox{\scriptsize Least}}=38.69\%$, Table \ref{tab:same_scenario_comparison}). 

We then test the models on the \texttt{Similar} scenario using the first half of AS Roma vs. Juventus FC as training set and the first half of match AS Roma vs. SS Lazio as test set. 
Note that this match is played by one of the teams that played the match used for training the model (AS Roma), and it is played in the same stadium (Stadio Olimpico, in Rome) and with the same light conditions (in the evening).
{\sf PassNet} outperforms {\sf ResBi} and the baselines ($F1_{\mbox{\scriptsize PassNet}}=61.74\%$, $F1_{\mbox{\scriptsize ResBi}}=59.34\%$), with an improvement of 20.19\% absolute and 48\% relative w.r.t. \texttt{LeastFrequent} ($F1_{\mbox{\scriptsize Least}}= 41.55\%$, Table \ref{tab:mixed_and_similar_scenarios_comparison}, right). 

\begin{figure*}[htb!]
\centering
    \subfigure[]{\includegraphics[width=0.44\linewidth]{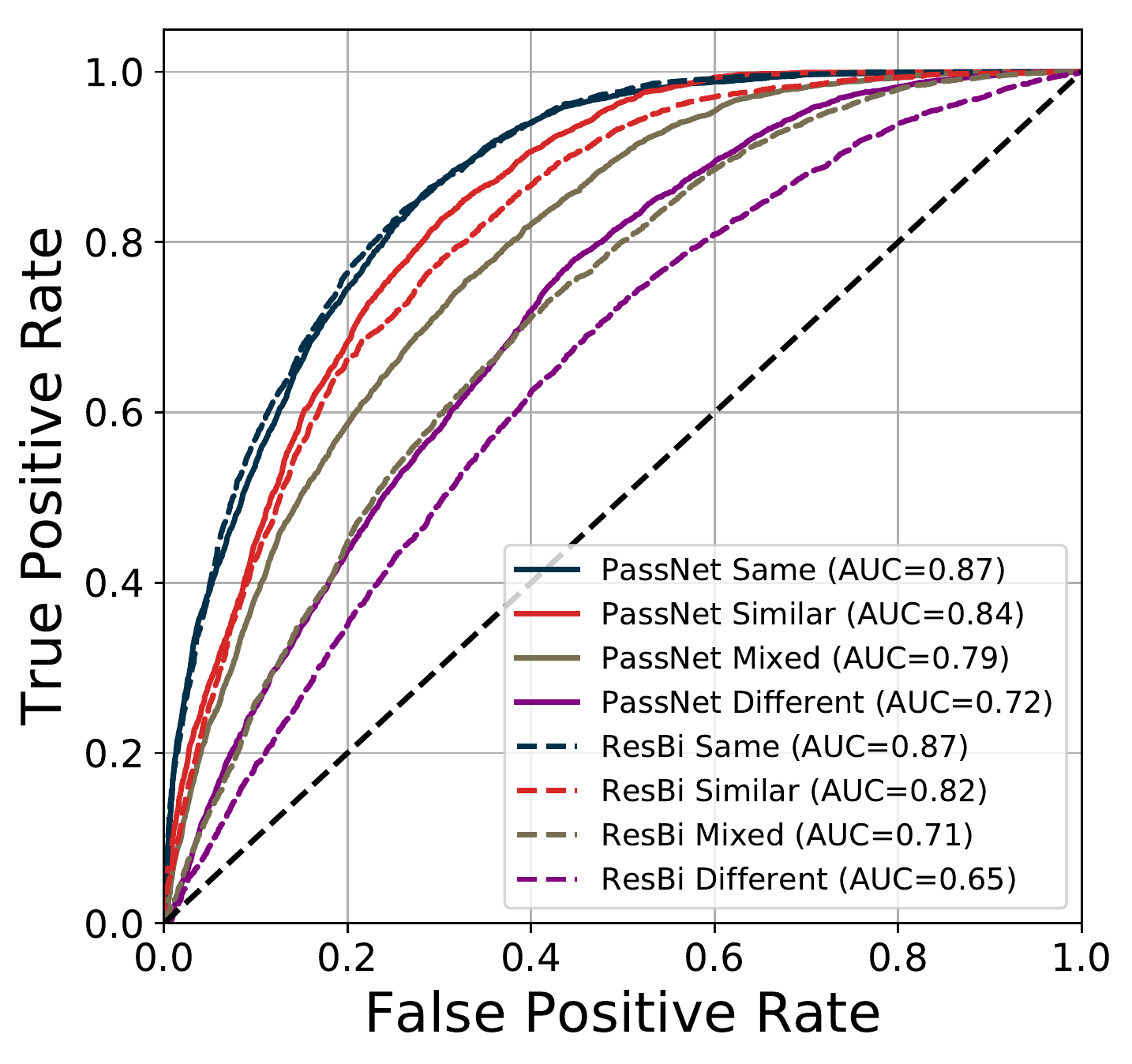}}
    \subfigure[]{\includegraphics[width=0.49\linewidth]{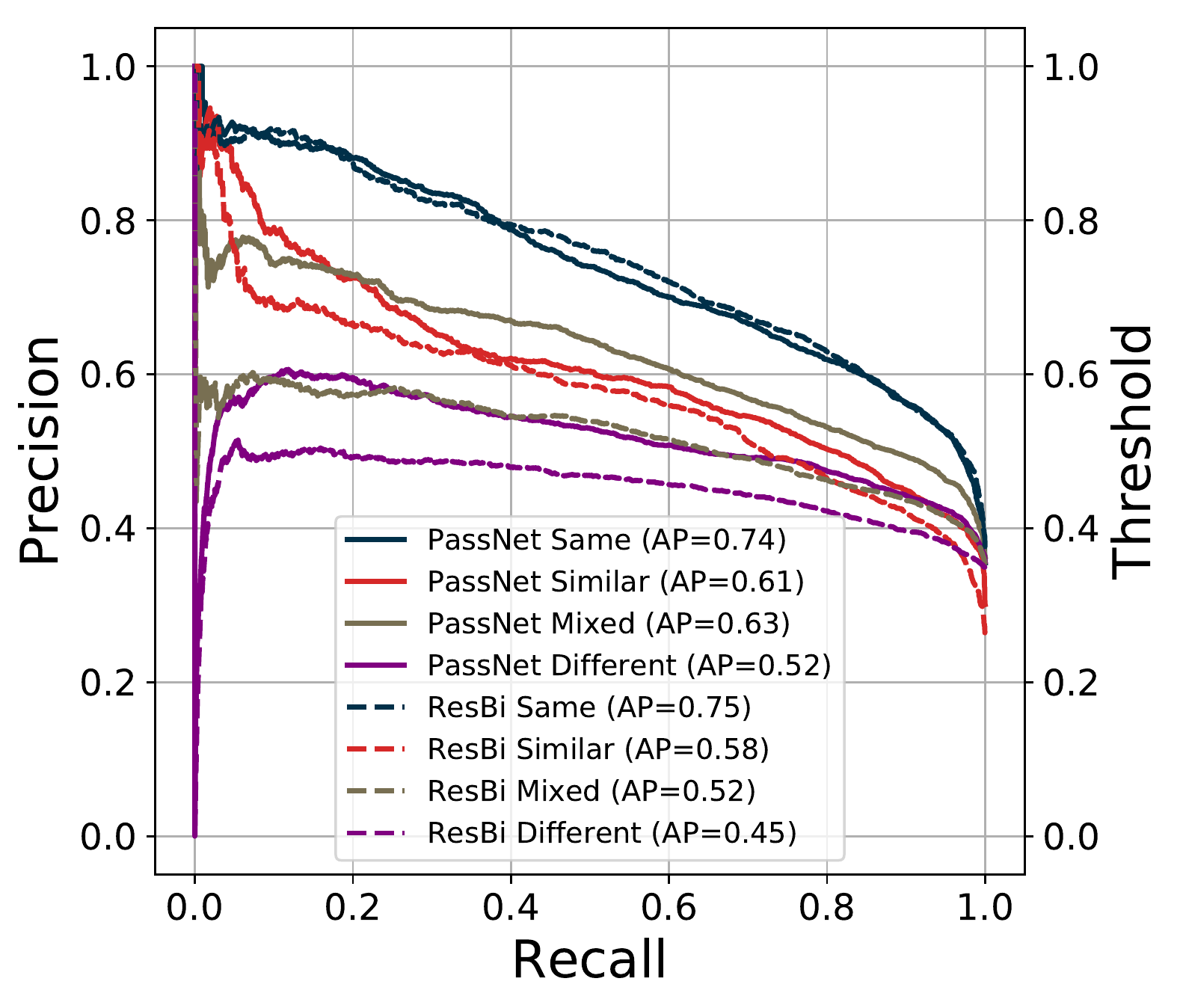}}
\caption{\textbf{Classification Performance.} (a) ROC curves and (b) precision and recall vayring the threshold, for {\sf PassNet} and {\sf ResBi} on the \texttt{Same} (AS Roma vs. Juventus FC), \texttt{Similar}, \texttt{Mixed} and \texttt{Different} (AC Chievo vs. Juventus FC) scenarios.}
\label{fig:roc_and_thresholds}
\end{figure*}

{\sf PassNet} outperforms all models on the \texttt{Mixed} scenario, too. Here we train the models using the first halves of AS Roma vs. Juventus FC and US Sassuolo vs. FC Internazionale and test them on AC Chievo Verona vs. Juventus FC. We obtain $F1_{\mbox{\scriptsize PassNet}}=63.73\%$ and $F1_{\mbox{\scriptsize ResBi}}=58.17\%$, with an improvement of 11.96\% absolute and 23\% relative with respect to \texttt{LeastFrequent} ($F1_{\mbox{\scriptsize Least}}=51.77\%$, see Table \ref{tab:mixed_and_similar_scenarios_comparison}, left). 
Finally, we challenge the models on the \texttt{Different} scenario, in which we use match AS Roma vs. Juventus FC to train the models and we test them on matches AC Chievo vs. Juventus FC and US Sassuolo vs. FC Internazionale. 
{\sf PassNet} and {\sf ResBi} have similar performance in terms of F1-score (Table \ref{tab:different_scenario_comparison}) and they both outperform \texttt{LeastFrequent}. Figure \ref{fig:roc_and_thresholds}a compares the ROC curves of {\sf PassNet} and {\sf ResBi} on the four experimental scenarios. 

\begin{table}[]
\newcolumntype{R}{>{\raggedleft\arraybackslash}X}
\begin{tabularx}{\linewidth}{l*{5}{R}}
\toprule                       & \bf scenario & YI & TH=.5 & TH=.9 \\ 
\cmidrule(l){2-5}
{\bf {\scriptsize AS} Roma vs. Juventus {\scriptsize FC}} 2H  & \texttt{Same}                     & \cellcolor{gray!8} $70.21$ {\scriptsize(.19)}           & $69.64$                      & $65.20$                      \\ 

{\bf {\scriptsize US} Sassuolo vs. {\scriptsize FC} Inter} 2H & \texttt{Same}                     & \cellcolor{gray!8} $54.44$ {\scriptsize  (.0005)}         & $40.60$                      & $28.26$                      \\ 
{\bf {\scriptsize US} Chievo vs. Juventus {\scriptsize FC}} 1H & \texttt{Mixed}                    & \cellcolor{gray!8} $63.73$ {\scriptsize(.003)}          & $42.55$                      & $27.22$                      \\ 
{\bf {\scriptsize AS} Roma vs. {\scriptsize SS} Lazio} 1H    & \texttt{Similar}                  & \cellcolor{gray!8} $61.74$ {\scriptsize(.15)}           & $61.70$                      & $55.24$                      \\
{\bf {\scriptsize US} Sassuolo vs. {\scriptsize FC} Inter} 2H & \texttt{Different}                & \cellcolor{gray!8} $53.92$ {\scriptsize(.004)}                   & $47.00$                           & $34.62$                           \\ 
{\bf {\scriptsize US} Chievo vs. Juventus {\scriptsize FC}} 1H & \texttt{Different}                & \cellcolor{gray!8} $59.51$ {\scriptsize(.0003)}                   & $23.17$                           & $8.45$                           \\ \bottomrule
\end{tabularx}
\caption{F1-score (in percentage) of {\sf PassNet} at different thresholds for each match/scenario. The best value for each combination of metric and threshold is highlighted in grey. For YI, we specify the value of the threshold in parenthesis.}
\label{tab:thresholds}
\end{table}

\begin{table}[htb!]
\begin{tabularx}{\linewidth}{l*{5}{r}X*{5}{r}}
\toprule
         & \multicolumn{5}{c}{\scriptsize \textbf{2H of AS Roma vs. Juventus FC}} && \multicolumn{5}{c}{\scriptsize \textbf{2H of US Sassuolo vs. FC Inter}} \\ \cmidrule(l){2-6} \cmidrule(l){7-12}
         & {\sf PassNet} & {\sf ResBi} & \multicolumn{3}{c}{\bf baselines} && {\sf PassNet} & {\sf ResBi} & \multicolumn{3}{c}{\bf baselines}\\ \cmidrule(l){4-6} \cmidrule(l){10-12}
         & \scriptsize YI${=}.19$ & \scriptsize YI${=}.05$ & \texttt{Random} & \texttt{Most} & \texttt{Least} && \scriptsize YI${=}.0005$ & \scriptsize YI${=}.01$ & \texttt{Random} & \texttt{Most} & \texttt{Least} \\ \midrule
$Acc$    &       76.07 &       77.18 &           50.09 &         67.32 &          32.68 &&       63.74 &       67.88 &           50.61 &         76.02 &          23.98 \\
\rowcolor[HTML]{EFEFEF} 
$F1$     &       70.21 &       70.50 &           39.74 &           0.0 &          49.26 &&       54.44 &       53.72 &           32.81 &           0.0 &          38.69 \\
$Prec$   &       59.17 &       61.03 &           32.82 &           0.0 &          32.68 &&       38.96 &       41.04 &           24.35 &           0.0 &          23.98 \\
$Rec$    &       86.32 &       83.45 &           50.36 &           0.0 &            100 &&       90.34 &       77.73 &           50.29 &           0.0 &            100 \\
$PrecNo$ &       91.45 &       90.22 &           67.46 &         67.32 &            0.0 &&       94.78 &       90.21 &           76.38 &         76.02 &            0.0 \\
$RecNo$  &       71.09 &       74.13 &           49.95 &           100 &            0.0 &&       55.34 &       64.77 &           50.70 &           100 &            0.0 \\
\bottomrule
\end{tabularx}
\caption{Comparison of {\sf PassNet}, {\sf ResBi} and the baselines (\texttt{Random}, \texttt{Most}, \texttt{Least}), on matches AS Roma vs. Juventus FC (left) and US Sassuolo vs. FC Internazionale (right), of the \texttt{Same} scenario. The metrics are specified in percentage. YI = Youden Index.}
\label{tab:same_scenario_comparison}
\end{table}

\begin{table}[htb!]
\begin{tabularx}{\linewidth}{X*{5}{r}c*{5}{r}}
\toprule
        &    \multicolumn{5}{c}{\scriptsize \textbf{1H of AC Chievo vs. Juventus FC}} &&    \multicolumn{5}{c}{\scriptsize \textbf{1H of AS Roma vs. SS Lazio}} \\ \cmidrule(l){2-6} \cmidrule(l){8-12}
        & {\sf PassNet} & {\sf ResBi} & \multicolumn{3}{c}{\bf baselines} && {\sf PassNet} & {\sf ResBi} & \multicolumn{3}{c}{\bf baselines} \\ \cmidrule(l){4-6} \cmidrule(l){10-12}
        & \scriptsize {YI=$.003$} & \scriptsize {YI=$.0064$} & \texttt{Random} & \texttt{Most} & \texttt{Least} && \scriptsize {YI=$.15$} & \scriptsize {YI=$.07$} & \texttt{Random} & \texttt{Most} & \texttt{Least} \\ \midrule
$Acc$   & 70.00          &    63.37     & 49.99        & 65.08         & 34.92          &&  73.13 &       72.32 &           50.30 &         73.78 &          26.22           \\
\rowcolor[HTML]{EFEFEF}
$F1$    & 63.73       &    58.17     & 39.59        & 0.0           & 51.77          && 61.74 &       59.34 &           34.90 &           0.0 &          41.55           \\ 
$Prec$  & 55.15       &    48.38     & 32.82        & 0.0           & 34.92          &&  49.26 &       48.25 &           26.58 &           0.0 &          26.22           \\ 
$Rec$   & 75.48       &    72.95     & 49.89        & 0.0           & 100            &&  82.68 &       77.05 &           50.81 &           0.0 &            100             \\ 
$PrecNo$& 83.60       &    80.05     & 67.13        & 65.08         & 0.0            && 91.89 &       89.65 &           74.14 &         73.78 &            0.0             \\ 
$RecNo$ & 67.06       &    58.23     & 50.04        & 100           & 0.0            && 69.73 &       70.63 &           50.12 &           100 &            0.0           \\
\bottomrule
\end{tabularx}
\caption{Comparison of {\sf PassNet}, {\sf ResBi} and the baselines (\texttt{Random}, \texttt{Most}, \texttt{Least}), on matches AC Chievo vs. Juventus FC (left, \texttt{Mixed} scenario) and AS Roma vs. SS Lazio (right, \texttt{Similar} scenario). The metrics are specified in percentage. YI = Youden Index.}
\label{tab:mixed_and_similar_scenarios_comparison}
\end{table}

\begin{table}[htb!]
\begin{tabularx}{\linewidth}{l*{5}{r}X*{5}{r}}
\toprule
         & \multicolumn{5}{c}{\scriptsize \textbf{1H of AC Chievo vs. Juventus FC}} && \multicolumn{5}{c}{\scriptsize \textbf{2H of US Sassuolo vs. FC Inter}} \\ \cmidrule(l){2-6} \cmidrule(l){7-12}
         & {\sf PassNet} & {\sf ResBi} & \multicolumn{3}{c}{\bf baselines} && {\sf PassNet} & {\sf ResBi} & \multicolumn{3}{c}{\bf baselines}\\ \cmidrule(l){4-6} \cmidrule(l){10-12}
         & \scriptsize YI${=.0003}$ & \scriptsize ${.00001}$ & \texttt{Random} & \texttt{Most} & \texttt{Least} && \scriptsize YI${=.004}$ & \scriptsize ${.0005}$ & \texttt{Random} & \texttt{Most} & \texttt{Least} \\ \midrule
$Acc$   & 63.33          &    59.82     & 49.99        & 65.08         & 34.92          &&       66.25 &       67.13 &           50.61 &         76.02 &          23.98 \\
\rowcolor[HTML]{EFEFEF} 
$F1$    & 59.51       &    53.42     & 39.59        & 0.0           & 51.77          &&       53.91 &       54.08 &           32.81 &           0.0 &          38.69 \\
$Prec$  & 48.43       &    44.88     & 32.82        & 0.0           & 34.92          &&       40.08 &       40.66 &           24.35 &           0.0 &          23.98 \\
$Rec$   & 77.16       &    65.97     & 49.89        & 0.0           & 100            &&       82.34 &       80.71 &           50.29 &           0.0 &            100 \\
$PrecNo$& 82.02      &    75.58     & 67.13        & 65.08         & 0.0            &&       91.65 &       91.17 &           76.38 &         76.02 &            0.0 \\
$RecNo$ & 55.91       &     56.52     & 50.04        & 100           & 0.0            &&       61.17 &       62.85 &           50.70 &           100 &            0.0 \\
\bottomrule
\end{tabularx}
\caption{Comparison of {\sf PassNet}, {\sf ResBi} and the baselines (\texttt{Random}, \texttt{Most}, \texttt{Least}) on AC Chievo vs. Juventus FC (left) and US Sassuolo vs. FC Internazionale (right), of the \texttt{Different} scenario. The metrics are specified in percentage. YI = Youden Index.}
\label{tab:different_scenario_comparison}
\end{table}


In summary, {\sf PassNet} and {\sf ResBi} significantly outperform the baselines on all four scenarios, indicating that our approach is able to learn from data to annotate pass events. We find that {\sf PassNet} outperforms {\sf ResBi} on all scenarios but the \texttt{Different} one, on which the two models have similar performance in terms of F1-score. 
In particular, {\sf PassNet} achieves the best performance on the \texttt{Same} scenario ($AUC=0.87$), followed by the \texttt{Similar} ($AUC=0.84$), the \texttt{Mixed} ($AUC=0.79$), and the \texttt{Different} ($AUC=0.72$) scenarios (Figure \ref{fig:roc_and_thresholds}a).

In general, our results highlight that the detection of the ball and the closest players to it makes a significant contribution to pass detection. 
This is not surprising on the \texttt{Same} scenario, in which we use the second half of the match the first half of which is used for training. 
However, the fact that the performance on the \texttt{Similar} scenario is better than the performance on the \texttt{Mixed} and the \texttt{Different} scenarios suggests that, provided the availability of matches for an entire season, we may build different models for different teams, for different light conditions, or a combination of both. 
Moreover, the similar performance of {\sf PassNet} and {\sf ResBi} on the \texttt{Different} scenario suggests that the contribution of the object detection module is weaker for matches whose video conditions differ significantly from those of the match used for training the model.
Note that the threshold with the maximum Youden Index provides also the best results in terms of F1-score (Table \ref{tab:thresholds}).

Figure \ref{fig:roc_and_thresholds}b shows how recall and precision change varying the threshold used to construct the passing vector, a useful tool for possible users of {\sf PassNet}, such as data collection companies: if they want to optimize the precision of pass detection, the plot indicates that high thresholds must be used. 
In contrast, if they want to optimize recall, thresholds in the range $[0.0, 0.4]$ must be preferred.


\begin{figure}[htb!]
\centering
    \includegraphics[width=0.9\linewidth]{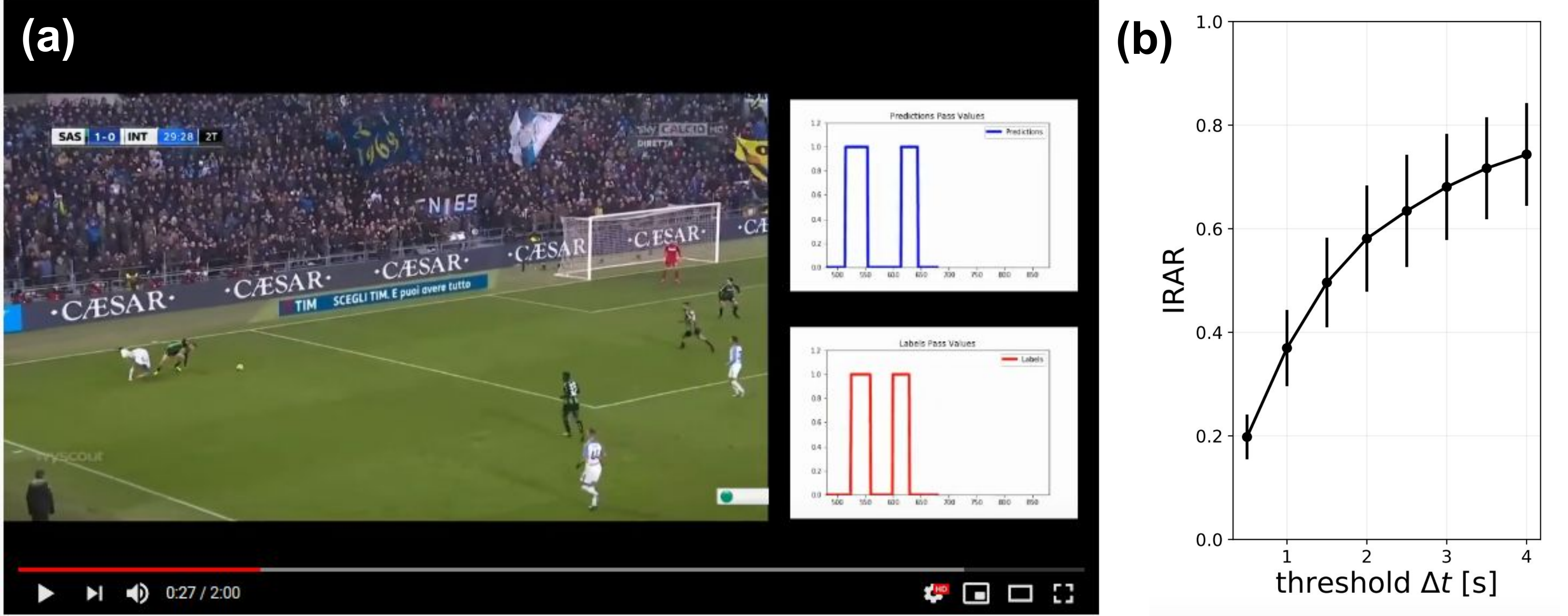}
\caption{\textbf{{\sf PassNet} in action.} (a) Structure of the video showing how {\sf PassNet} annotates US Sassuolo vs. FC Internazionale as the match goes by. The left side shows the match, a label ``Pass'' appears every time a pass is detected. The right side shows two animated plots comparing the real (red) and the predicted (blue) labels. (b) Average IRAR w.r.t. Wyscout operators varying the time threshold $\Delta t$.}
\label{results:6}
\end{figure}

To visualize the limits of {\sf PassNet}, we create some videos that show the results of its predictions as the match goes by. In particular, we make publicly available video clips for US Sassuolo vs. FC Internazionale (\texttt{Different} scenario) and AS Roma vs. Juventus FC (\texttt{Same} scenario).\footnote{\texttt{\url{https://youtu.be/l4Qt1tjfE_8}}, \texttt{\url{https://youtu.be/sOxYG4Fduoc}}}
Figure \ref{results:6}a shows the structure of these videos. On the left side, we show the match, in which a label ``Pass'' appears every time {\sf PassNet} detects a pass. On the right side, we show two animated plots that compare the real label with the model's prediction. In these plots, value 0 indicates no pass, value 1 indicates that there is a pass. 

The observation of these videos reveals that {\sf PassNet} sometimes classifies consecutive passes that come in a close interval of time as a single pass (AS Roma vs. Juventus FC). This is presumably because the YI threshold cannot detect the slightest changes that occur between two consecutive passes. 
Interestingly, the videos also reveal the presence of errors during the manual annotation. For example, in US Sassuolo vs. FC Internazionale, {\sf PassNet} recognizes passes that actually take place but that were not annotated by the human operators. Another error, that may constitute room for future improvement, is that {\sf PassNet} usually misclassifies as passes situations in which a player runs in possession of the ball.


As a further assessment of the reliability of the annotation made by our system, we evaluate the degree of agreement on matches in Tables \ref{tab:same_scenario_comparison}, \ref{tab:mixed_and_similar_scenarios_comparison}, and \ref{tab:different_scenario_comparison} between {\sf PassNet} and Wyscout's human operators computing the Inter-Rater Agreement Rate, defined as 
\begin{math}
\mbox{IRAR} {=} 1 - \frac{1-p_o}{1-p_e} \in [0, 1]
\end{math},
where $p_o$ is the relative agreement among operators (ratio of passes detected by both) and $p_e$ is the probability of chance agreement \cite{liu2013inter}.
In order to compute IRAR, we first associate each pass annotated by {\sf PassNet} at time $t$ with the Wyscout pass (if any) in the time interval $[t{-}\Delta t, t{+}\Delta t]$ (see \cite{liu2013inter,pappalardo2019playerank}).
Figure \ref{results:6}b shows how the mean IRAR varies with $\Delta t$: at $\Delta t {=} 1.5$s, mean $\mbox{IRAR}\approx0.50$, referred to as ``moderate agreement'' in \cite{viera2005understanding}, at $\Delta t {=} 3$s, mean $\mbox{IRAR}\approx0.70$, referred to as ``good agreement''. These results are promising considering that the typical agreement between two Wyscout human operators with $\Delta t = 1.5$ is $\mbox{IRAR}{=}0.70$ \cite{pappalardo2019playerank}. 

\section{Conclusion}
\label{sec:conclusion}
In this article, we presented {\sf PassNet}, a method for automatic pass detection from soccer video streams. 
We showed that {\sf PassNet} outperforms several baselines on four different scenarios, and that it has a moderate agreement with the sets of passes annotated by human operators. 

{\sf PassNet} can be improved and extended in several ways. 
First, in this article, we use broadcast videos that contain camera view changes and play-breaks such as replays, checks at the VAR, and goal celebrations. 
These elements may introduce noise and affect the performance of the model. 
The usage of fixed camera views and play-break detection models may be used to clean the video streams, reduce noise, and further improve the performance of {\sf PassNet}. 
Second, we use a pre-trained YOLOv3 that can recognize generic persons and balls. Although our results show that it provides a significant contribution to the predictions, we may build an object detection module to recognize specifically the soccer players and ball. 
Given the flexibility of YOLOv3's architecture, this may be achieved simply by training the model on a labeled data set of soccer players and balls. 
Moreover, we may integrate in {\sf PassNet} existing methods to detect the identity of players, for example by recognizing their jersey number.
Finally, {\sf PassNet} can be easily adapted to annotate other crucial events, such as shots, fouls, saves, and tackles, or to discriminate between accurate and inaccurate passes. 
Given the modularity of our model, this may be achieved simply by training the model using frames that describe the type of event of interest. 

In the meanwhile, {\sf PassNet} is a first step towards the construction of an automated event detection tool for soccer. 
On the one hand, this tool may reduce the time and cost of data collection, by providing a support to manual annotation. 
For example, event annotation may be partially delegated to the automatic tool and human annotators can focus on data quality control, especially for complex events such as duels and defending events.
On the other hand, it would consent to extend the data acquisition process to unexplored directions, such as matches in youth and non-professional leagues, to leagues far away in the past, and to practice matches in training sessions, allowing researchers to compare technical-tactical characteristics across divisions, times and phases of the season.

\section*{Acknowledgments}
This work has been supported by project H2020 SoBigData++ \#871042.
%
%
%

\bibliographystyle{splncs04}
\bibliography{biblio}
%

\end{document}